\documentclass[letterpaper,10pt,conference]{ieeeconf}
\IEEEoverridecommandlockouts
\usepackage{amsfonts,amsmath,amssymb,mathtools,bbm,bm}
\usepackage{diagbox}
\usepackage{hyperref}
\usepackage{graphicx,caption,subcaption}
\usepackage{booktabs}
\usepackage[ruled, lined, longend, linesnumbered, noend]{algorithm2e}

\usepackage[normalem]{ulem}

\newdimen{\algindent}
\usepackage{cite}
\usepackage{xcolor}

\newcommand\redout{\bgroup\markoverwith
{\textcolor{red}{\rule[0.5ex]{2pt}{0.8pt}}}\ULon}

\title{\LARGE \bf
Multi-agent Path Finding for Mixed Autonomy Traffic Coordination}

\author{Han Zheng, Zhongxia Yan, Cathy Wu 
\thanks{This work was supported by the National Science Foundation (NSF) CAREER award 2239566 and the MIT Amazon Science Hub.}
\thanks{Han Zheng, Zhongxia Yan and Cathy Wu are with the Laboratory for Information \& Decision Systems (LIDS), Massachusetts Institute of Technology, Cambridge, MA 02139, USA. Email:\{{\tt\small hanzheng,zxyan, cathywu\}@mit.edu}}
}

\begin{document}

\maketitle
\thispagestyle{empty}
\pagestyle{empty}

\begin{abstract}
In the evolving landscape of urban mobility, the prospective integration of Connected and Automated Vehicles (CAVs) with Human-Driven Vehicles (HDVs) presents a complex array of challenges and opportunities for autonomous driving systems. While recent advancements in robotics have yielded  Multi-Agent Path Finding (MAPF) algorithms tailored for agent coordination task characterized by simplified kinematics and complete control over agent behaviors, these solutions are inapplicable in mixed-traffic environments where uncontrollable HDVs must coexist and interact with CAVs. Addressing this gap, we propose the Behavior Prediction Kinematic Priority Based Search (BK-PBS), which leverages an offline-trained conditional prediction model to forecast HDV responses to CAV maneuvers, integrating these insights into a Priority Based Search (PBS)  where the A* search proceeds over motion primitives to accommodate kinematic constraints. We compare BK-PBS with CAV planning algorithms derived by rule-based car-following models, and reinforcement learning. Through comprehensive simulation on a highway merging scenario across diverse scenarios of CAV penetration rate and traffic density, BK-PBS outperforms these baselines in reducing collision rates and enhancing system-level travel delay. Our work is directly applicable to many scenarios of multi-human multi-robot coordination.
\end{abstract}

\section{Introduction}
\label{sec:introduction}
The rapid advancements in the field of autonomous vehicles and intelligent transportation systems herald a future where Connected and Automated Vehicles (CAVs) could become an integral part of our urban landscapes. This integration promises to significantly enhance traffic efficiency, reduce congestion, and improve overall road safety. However, the transition towards fully automated traffic systems is fraught with challenges, particularly due to the necessity for CAVs to coexist and interact seamlessly with Human-Driven Vehicles (HDVs) in mixed-traffic environments. The unpredictable nature of human driving behavior introduces a level of complexity that current autonomous driving systems must adeptly navigate to ensure harmonious and safe traffic flow.

Recent works in robotics, planning, and artificial intelligence have made significant strides in Multi-Agent Path Finding (MAPF) algorithms, which are designed to optimize multi-agent coordination. These algorithms have been instrumental in solving the coordination problems in various domains, including warehouse automation, drone swarms, and full-autonomy traffic coordination. However, their application in mixed-traffic scenarios, where autonomous systems must interact with uncertain human drivers, reveals significant limitations.

The core challenge addressed in this article lies in the assumptions of complete control over agent behaviors, which does not hold in road traffic scenarios involving HDVs.
We thereby introduce a novel algorithm called Behavior Prediction Kinematic Priority Based Search (BK-PBS) with two purposes: 1) predict the future behavior of HDVs and enable CAVs to avoid collisions with HDVs, 2) indirectly control and coordinate the behavior of HDVs through proactive interactions between CAVs and HDVs. The original Priority Based Search (PBS) algorithm \cite{ma2019searching} plans independent shortest paths for all agents, then resolves conflicts by searching over assignments of agent priorities and replanning agents to avoid conflicts with higher-priority agents. In BK-PBS, we enhance this approach by first training an offline conditional prediction model that forecasts HDV reactions to CAV maneuvers. This model is then invoked when a CAV is assigned a higher priority than an HDV, allowing PBS to reason about forecasted HDV trajectories conditioned on nearby CAVs' trajectories. This method fosters a proactive coordination strategy between CAVs and HDVs, moving beyond the reactive coordination of treating HDVs as obstacles.

To replan CAVs within BK-PBS, our low-level A* search utilizes motion primitives to accommodate kinematic constraints characterizing the realistic motion of CAVs. We demonstrate the effectiveness of BK-PBS in a simulated mixed-traffic scenario where CAVs and HDVs drive along a highway with merging lanes. Experimental result shows that BK-PBS outperforms baselines in reducing collision rates and system-level travel delay under different CAVs penetration rate and traffic density.

\begin{figure*}[t!]
\centering
\includegraphics[width=\textwidth]{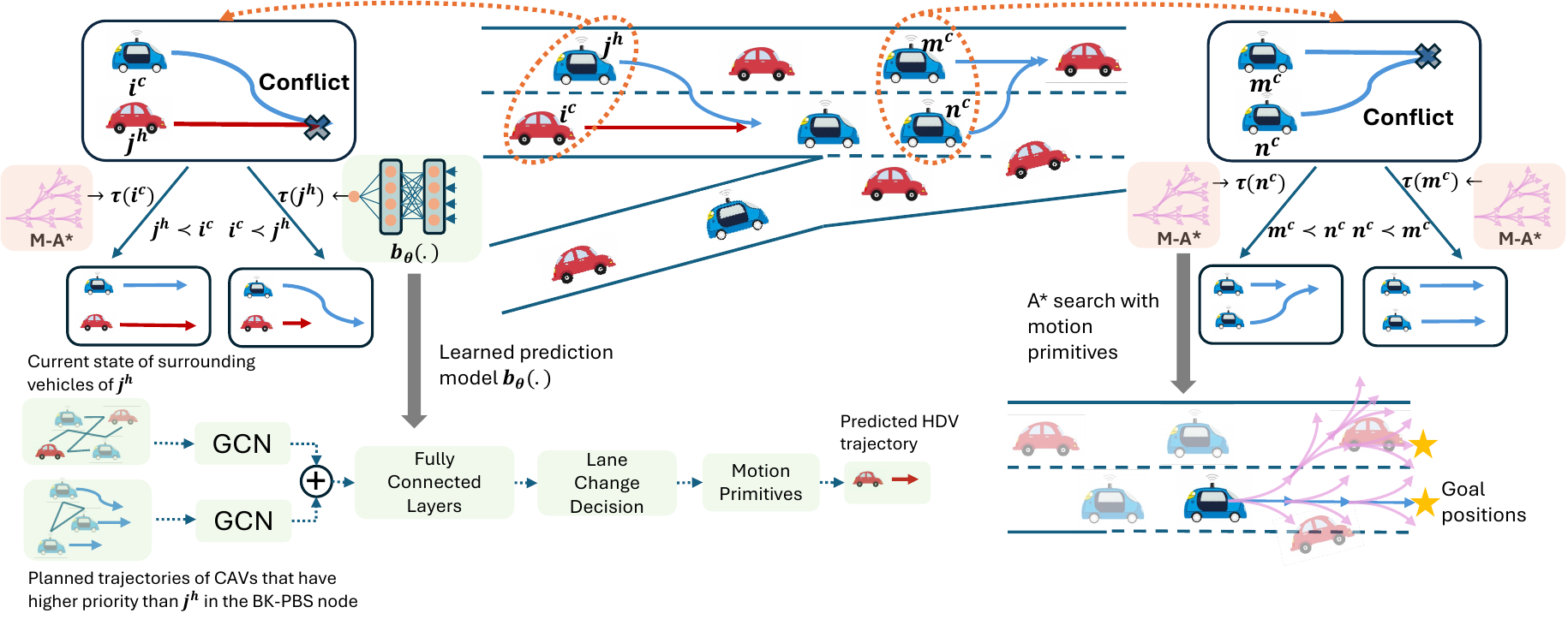}
\caption{Mixed traffic coordination: 1) The left portion describes a collision between a CAV $i^c$ and a HDV $j^h$. To resolve this CAV-HDV collision, BK-PBS introduces priority between these two vehicles, and replan the CAV trajectory using M-A* while re-predict the HDV trajectory using the conditional prediction model $b_{\theta}$. 2) The right portion describes a collision between a CAV $m^c$ and a CAV $n^c$. To resolve this CAV-CAV collision, BK-PBS introduces priority between these two vehicles, and replan each of the CAV trajectory using M-A*.}
\label{fig:1}
\end{figure*}

\section{Related Work}
\label{sec:related_work}

\subsection{Mixed-traffic Coordination}
While numerous works have proposed coordination methods for CAVs in various traffic settings \cite{rios2016survey}, fewer works have studied mixed autonomy settings where CAVs must coexist with HDVs in the same traffic system. Previous model-based optimal control methods for mixed autonomy typically design algorithms for smoother car-following, platooning, and other relatively simple behaviors \cite{keqiang2022, jiawei2021}. Other works have leveraged model-free reinforcement learning (RL) \cite{wu2021flow, yan2022unified} to automatically discover high-performing control policies for a variety of traffic scenarios, though without explicit collision prevention. Nevertheless, we are the first to study a search-based planning method for mixed autonomy, and we additionally incorporate behavior prediction of surrounding HDVs rather than assume their behavior is fully characterized.

\subsection{Behavior Prediction}

Behavior and trajectory prediction constitutes a cornerstone of research in the domain of autonomous driving, offering substantial contributions to the understanding and anticipation of the movements of pedestrians and human-operated vehicles in the vicinity of autonomous vehicles \cite{Mozaffari2022}. This field has traditionally favored deep learning approaches for forecasting future positions and actions\cite{pmlr-v155-zhao21b, Ivanovic2018TheTP, Gupta_2018_CVPR}.

However, the scope of this paper diverges from the task of designing a deep learning model for behavior and trajectory prediction based on extensive real-world datasets. Instead, our research centers on the integration of an existing prediction model into the strategic planning processes of a fleet of autonomous vehicles operating within mixed traffic environments. Our primary interest lies in exploring the synthesis of predictive insights with the dynamic decision-making required for effective navigation and safety in scenarios where CAVs and HDVs coexist.

\subsection{Multi-agent Path Finding}
The classical multi-agent path finding (MAPF) problem \cite{stern2019multi} is a NP-hard \cite{yu2013structure} problem which seeks to find the shortest collision-avoiding paths for a set of agents in a discrete graph. Since the space of joint agent trajectories is intractably large to consider \cite{sharon2015conflict}, nearly all MAPF algorithms rely on repeatedly calling a single-agent path planner such as A* search \cite{hart1968formal} or SIPP \cite{phillips2011sipp}, while holding paths of some set of other agents as constraints. In particular, Conflict-based Search (CBS) \cite{sharon2015conflict} is a seminal algorithm which first plans independent shortest paths for all agents then relies on backtracking tree-search to resolve pairwise agent collisions by adding local constraints. Priority-based Search (PBS) \cite{ma2019searching} is a scalable extension of CBS, albeit suboptimal and incomplete, which searches over global priorities between conflicting agents rather than local constraints.

In continuous space, several works in multi-robot motion planning \cite{kottinger2022conflict,okumura2022quick} have applied sampling based methods like probabilistic roadmaps \cite{kavraki1996probabilistic} in conjunction with CBS or other discrete MAPF algorithms. to plan over settings with continuous 2D space and time. Alternatively, continuous settings may be discretized for application of MAPF algorithms \cite{honig2018trajectory}. \cite{yan2024multi} proposes a MAPF-based approach for coordination of CAVs in a full autonomy system, but no MAPF-based approach has ever been designed for mixed autonomy to our knowledge.

\section{Problem Formulation}
\label{sec:formulation}
\subsection{MAPF Formulation of Highway Coordination}
\label{subsec: MAPF Formulation}

We formulate the highway coordination problem in the Multi-Agent Path Finding (MAPF) framework. The classical MAPF problem involves $k$ agents and is defined by the tuple $\langle G, s, g \rangle$, where $G = (V, E)$ represents an undirected graph. In this context, $s: [1, \ldots, k] \rightarrow V$ assigns each agent to a source position, while $g: [1, \ldots, k] \rightarrow V$ assigns each agent to a goal position. We focus on discrete time domain, with each agent occupying a position on the graph at any given timestep and capable of performing one action that determines the next position node to transit to, per timestep. 

Applying this framework to a mixed-traffic highway scenario, we represent each CAV agent in the scene as $i^{c} \in \mathbf{C}$, and each HDV agent as $i^{h} \in \mathbf{H}$. In coordinating highway traffic, our primary objective is to determine the control inputs for the CAVs; therefore, we only set goals for CAV agents. We operate within a discrete-time approximation of a continuous control task, and we solve the planning problem using motion primitives in a rolling horizon fashion. 

Let the position of a CAV agent at time $t$ be $\mathbf{x}_t(i^{c}) \in \mathbb{R}^2$, representing the source position. As CAVs are generally expected to drive forward or merge based on the control inputs, their goals are not restricted to a single position but rather to a set of positions. We then define the goal set at time $t$ for a CAV agent as $\mathbf{G}_t(i^c)$. If the CAV is traveling on the main road, its goal set $\mathbf{G}_t(i^c)$ comprises positions on its current and adjacent lanes that lie ahead at or farther than distance $d$. Conversely, for a CAV on a merging lane, the goal set $\mathbf{G}_t(i^c)$ includes positions on any adjacent lane before the merging zone ends.

In the classical MAPF context, actions are determined by a function $u: V \rightarrow V$, where $u(v) = v'$ indicates that an agent at position $v$ will move to position $v'$ in the following timestep. In our highway coordination task, we introduce motion primitives as actions to address the continuous control challenge. These motion primitives encompass predefined high-level maneuvers such as acceleration and lane changing. For an action $u^c \in \mathbf{U}^c$ drawn from the motion primitive library, a CAV agent progresses to the subsequent position in accordance with a transition model $f^c$. The transition between CAV positions is expressed as $\mathbf{x}_{t+1}(i^{c}) = f^c(\mathbf{x}_t(i^{c}), u^c)$. The movement of a HDV is not controlled by our planning algorithm but in react to its surrounding vehicles. It follows $\mathbf{x}_{t+1}(j^{h}) = f^h(\mathbf{x}_t(j^{h}), \mathbf{S}_t(j^h))$, where $f^h$ is the HDV transition model and $\mathbf{S}_t(j^h)$ is the collection of vehicles states (position, velocity and acceleration) of vehicles in the vicinity of HDV $j^h$.

We denote the sequence of CAV actions $h_i^c = [u_0^c, \ldots, u_T^c]$ within a control horizon $T$ pertaining to CAV agent $i^{c}$ as a plan, and we define the trajectory sequence associated with this plan as $\tau (i^{c}) = [\boldsymbol{x}_{1}(i^{c}), \ldots, \boldsymbol{x}_{T+1}(i^{c})]$. A solution to the problem is thus defined as a collection of $|\boldsymbol{C}|$ single-agent CAV plans, one for each CAV, ensuring that all CAVs reach their respective goals from their sources without colliding with any other vehicles on road. 

The collision free constraints can be represent as  $\forall t \in \{1,\ldots, T+1\}, \forall j \in \boldsymbol{H}\cup \{\boldsymbol{C}/{i}^{c}\}, \boldsymbol{B}_t(i^{c}) \cap \boldsymbol{B}_t(j) = \emptyset$, where $\boldsymbol{B}_t(\cdot)$ denotes the bounding box of a vehicle at time $t$. The center of a bounding box is the vehicle position.

 Beside the collision-free constraint, we also aim to facilitate smooth merging and enhance the highway's throughput. This objective can be represented as minimizing the total travel delay cost 
 \begin{equation}
 \label{eq:delay}
    J = \sum_{i\in \boldsymbol{H}\cup \boldsymbol{C} } [t^f_i - t^{*}_i]
 \end{equation}
  where $t^{*}_i$ is the minimal time that vehicle $i$ arrives at the end of the highway section, and $t_i^f$ is the actual arrival time after planning. The free-flow travel time $t^{*}$ can be computed under the assumption that the vehicle accelerates and travels at maximum speed ignoring other on-road vehicles.

\subsection{Kinematic Bicycle Model}
\label{bicycle_model}
We adopt the bicycle model \cite{rajamani2011vehicle} for the vehicle kinematics, and the motion primitives translate into low-level bicycle kinematics. The bicycle model allows for a more accurate representation of real-world vehicle motion compared to point-mass models while remaining computationally efficient for trajectory planning and control. Here, the pairs of front and rear wheels are each aggregated into a singular wheel along the midline of the vehicle. The control inputs are acceleration $a$ and front wheel steering angle $\delta$. The distance from the center of gravity to front and rear wheels is half of vehicle length $\ell_k$. $\psi$ denotes the heading. $\beta$ denotes the slip angle.

\begin{multline}
\dot{x}_x=v \cos (\psi+\beta) \hspace{1cm} \dot{x}_y=v \sin (\psi+\beta) \hspace{1cm} \dot{v} = a \\
 \dot{\psi}=\frac{v \cos (\beta)}{\ell_k}\tan \left(\delta\right) \hspace{1cm} \beta=\tan^{-1}\left(\frac{\tan \delta}{2}\right)\\
\label{bm_eq}
\end{multline}

\section{Method}
\label{sec:method}
For coordinating mixed-traffic, we take insight from a MAPF algorithm called Priority Based Search (PBS) and incorporate an offline learned prediction model to predict and revise the future behavior of HDVs in the PBS planning of CAV trajectories. To accommodate the constraints imposed by the vehicle kinematic model, a multi-phase A* search searches on a discrete state graph where the edges are represented by motion primitives. We name our proposed method as Behavior Prediction Kinematic Priority Based Search (BK-PBS), which is sketched in Algorithm~\ref{alg:BK-PBS}.

\subsection{Multi-phase A* with Motion Primitives}
\label{subsec:A_star}
 In the context of classical MAPF, system dynamics are often reduced to simplified point-mass agents, with low-level path-finding such as A* search conducted in a grid-based world devoid of kinematic constraints. For the highway coordination problem, we aim to construct a continuous trajectory for each CAV from its starting point to its destination, in compliant with the vehicle kinematics detailed in the referenced bicycle model, while avoiding collisions with other vehicles.

To address this, we present a method that combines kinematic constraints with A* search — Multi-phase A* with Motion Primitives (M-A*). Let $i^{c}$ be the CAV under planning  associated with a BK-PBS node $N$. This method performs the A* search on a graph where each node represents the position of $i^{c}$, and each edge represents a motion primitive — a viable maneuver that conforms to the kinematic constraints outlined by the bicycle model. Trajectories of vehicles other than the CAV under planning are integrated into the graph as dynamic obstacles constraints. 

Each motion primitive delineates a possible segment of  $\tau(i^{c})$, distinguished by particular steering and velocity profiles. Given a target position of $i^{c}$, the A* algorithm searches for the sequence of primitives that leads to the lowest cost, collision-free trajectory from the current state to the target. The motion primitive library includes the following maneuvers: 1) accelerate, 2) decelerate, 3) idle, 4) lane change left, 5) lane change right, and 6) emergency braking.

For the feasibility of low-level planning, we enhance the A* search with a tiered planning approach. In the initial planning phase, A* search endeavors to locate the lowest-cost, collision-free path from the current location of $i^{c}$ to its goal location within a specified search budget. The search biases towards forward movement, with a larger share of the budget dedicated to acceleration and lane-changing primitives. If the initial search does not yield a valid path, a secondary search is conducted that biases towards safe deceleration. We restrict the primitive library to exclude the acceleration and idle maneuvers. For CAVs driving on the main road, we revise its goal to closer positions on its current and adjacent lanes that lie ahead at distance $d'$, which is smaller than the original goal distance $d$. We also adjust the budget's distribution, favoring deceleration and emergency braking. Subsequently, a second phase of search commences to determine a more conservative, collision-free trajectory.

\subsection{Conditional Prediction Model}
\label{sec:Prediction }

In this study, we address the challenge of coordinating a mixed traffic system comprising both HDVs, which exhibit uncertain and uncontrollable behavior, and CAVs. The uncertainty of HDVs complicates the application of PBS for traffic planning. Specifically, if a CAV cannot anticipate an HDV's future movements, planning a collision-free trajectory for the CAV becomes infeasible. A potential solution involves predicting the trajectories of HDVs in advance, treating them as dynamic obstacles in the CAVs' path, and utilizing PBS to navigate CAVs around HDVs. This approach, which we refer to as reactive coordination, inherently prioritizes HDVs within the PBS framework, compelling CAVs to adjust based on HDV actions. However, our objective is to establish a proactive coordination strategy, wherein HDV behavior is also influenced by CAV movements. This is achieved by branching over the priority order between CAVs and HDVs in conflict situations within PBS. Nonetheless, the challenge arises in situations where an HDV is assigned a lower priority than a CAV, as the HDV's trajectory cannot be directly controlled or replanned.

To address this issue, we introduce a conditional prediction model, denoted as $b_{\theta}$, designed to predict the future actions of HDVs based on potential CAV maneuvers. This model enables us to plan for CAV movements in such a way that avoids collisions with HDVs, taking into account the predicted responses of HDVs to CAV actions. Essentially, this model allows for the adjustment of HDV priorities and the subsequent reevaluation of their trajectories in light of CAV movements, facilitating proactive coordination.

We illustrate PBS's branching between a CAV and a HDV in the left part of Fig.\ref{fig:1}.

\subsection{Mixed-traffic Priority Based Planning}
\label{sec:PBS}
BK-PBS is a centralized planning algorithm that determines actions for all CAVs at once. Like PBS, BK-PBS is a depth-first traversal of a Priority Tree (PT). Let $i \prec j$ denotes that vehicle $i$ has priority over vehicle $j$. Each node $N$ encompasses a priority ordering among vehicle pairs $(i, j)$ through the priority order $\prec_N $ associated with node $N$, along with a set of minimum-cost trajectories that comply with this priority ordering, ensuring no collision between any vehicle trajectory under the given priority constraints. Given an agent $i$, we denote the set of agents that has higher or lower priority than $i$ in node $N$ as $\{\prec_N i\}$ or $\{\succ_N i\}$, respectively. The objective of minimizing total travel delay cost $J$ is equivalent to maximize the overall travel speed. 

The root node of the PT contains no predefined priorities (line 2) and the set of independent minimum-cost trajectories for all vehicles (line 3-7). For a CAV, the root node trajectory is found using M-A* without considering the trajectory of other vehicles as dynamic obstacle constraints; while a HDVs, the root node trajectory is predicted by $b_{\theta}$ without considering any CAV actions.

When the search arrives at a node $N$ in the PT, if no collision are detected among vehicles, $N$ is identified as a goal node in the high-level search (differing from the \textit{goal locations} described in \ref{sec:formulation}). We assume that two HDVs are capable of avoiding collisions with one another through natural coordination, thus eliminating the need for branching in conflicts involving solely HDVs. If a collision arises between a CAV $i$ and any vehicle $j$ (line 14), BK-PBS branches $N$ into two new child nodes, respectively introducing new priority constraints $i \prec j$ and $j \prec i$ to the child nodes (line 17). The left section of Fig.\ref{fig:1} illustrates the branching of a CAV-HDV collision, while the right section shows the branching of a CAV-CAV collision. Within each of the newly generated child nodes, a replanning process is invoked to resolve collisions according to the priority ordering (line 18). Within the replanning, BK-PBS first conducts a topological sort to arrange the vehicle indices in accordance with the priority ordering (line 24). It then plans a conflict-free minimum-cost trajectory for each agent subject to the trajectories of all agents with higher priority.

For a CAV-HDV collision, assigning a higher priority to an HDV results in adding the predicted HDV trajectory as a constraint in the CAV's low-level planning via the M-A* (line 28). Conversely, if a lower priority is assigned to an HDV, $b_{\theta}$ is invoked to re-predict the HDV's trajectory, conditioned on the CAV's planned trajectory (line 31). It is important to note that within the BK-PBS update plan, the input for $b_{\theta}$ includes the trajectories of all CAVs with higher priority, ensuring the preservation of established priority orders. For a CAV-CAV collision, we use M-A* (line 28) to find the the trajectory for each CAV subject to the constraints induced by the new priority order. 

A child node is pruned if for some agent, no trajectory can be found to avoid conflict with all higher-priority agents (line 32). If no viable child node emerges from an expanded $N$, BK-PBS backtracks to the parent node. If two child nodes emerge from an expanded $N$, the child node with lower sum of cost will be searched first.

\begin{algorithm}[h!]
\caption{BK-PBS for mixed traffic coordination}\label{alg:BK-PBS}
\SetKwFunction{BKPBS}{BK-PBS}
\SetKwProg{Fn}{Function}{:}{}
\Fn{\BKPBS{$\boldsymbol{C}, \boldsymbol{H}, \prec_0$, \textcolor{black}{$b_{\theta}$}}}{{
$\prec_{root} \leftarrow \prec_0$; $Root.plan \leftarrow \emptyset$\;
\For{$i \in \boldsymbol{C} \cup \boldsymbol{H}$}{
    $success \leftarrow \texttt{UpdatePlan}(Root, i, b_{\theta})$\;
    \If{not $success$}{
        \Return ``No Solution''\;
    }
}
$Root.cost \leftarrow -\sum$ vehicle speeds in $Root.plan$\;
$STACK \leftarrow \{Root\}$\;
\While{$STACK \neq \emptyset$}{
    $N \leftarrow$ top node in $STACK$\;
    $STACK \leftarrow STACK \setminus \{N\}$\;
    \If{$N.plan$ has no collision}{
        \Return $N.plan$\;
    }
    $C \leftarrow$ first pair of CAV-involved collision $(i, j)$ in $N.plan$\;
    \For{$i$ in $C$}{
        $N' \leftarrow N$ \;
        $\prec_{N'} \leftarrow \prec_{N} \cup\ \{j \prec i\}$\;
        
        $success \leftarrow \texttt{UpdatePlan}(N', i, b_{\theta})$\;
        \If{$success$}{
            Add $N'$ into $STACK$ in non-increasing order of $N'.cost$\;
        }
    }
    
}
\Return ``No Solution''\;}
}\

\SetKwFunction{FUpdatePlan}{UpdatePlan}
\SetKwProg{Fn}{Function}{:}{}
\Fn{\FUpdatePlan{$N, i, b_{\theta}$}}{
    {$L \leftarrow$ topological ordering set $\{i\} \cup \{\succ_{N} i\}$\;
    \For{$q \in L$}{
        \If{$q = i$ or $\exists k \in \{\prec_{N} i\}$ that $q$ collide with $k$ in $N.plan$}{
        \eIf{$q \in \boldsymbol{C}$}{$\tau(q) \leftarrow \text{M-A*}(\{\tau(p) \mid p \prec_N q \})$\;}
        {$o(q) \leftarrow $ surrounding vehicle states\;
        predict the trajectory \textcolor{black}{$\tau(q) \leftarrow b_{\theta}(o(q),\{\tau(p) \mid p \prec_N q,\ p \in \boldsymbol{C}\})$}}
        
            \If{no $\tau(q)$ is returned by \textup{M-A*} or predicted $\tau(q)$ still lead to collision }{
                \Return false\;
            }
        update $\tau(q)$ in $N.plan$\;}
    }
    \Return true\;}
    
}

\end{algorithm}

\subsection{Neural Network-based Conditional Prediction Model}
\label{subsec: GCN}
We parameterize $b_{\theta}$ using neural networks. The architecture of this model is illustrated in the lower-left portion of Fig.\ref{fig:1}. Our approach utilizes Graph Convolutional  Networks (GCNs) \cite{KipfGCN} to represent the inter-vehicle dependencies.

Let $j^{h}$ be the HDV under prediction associated with a BK-PBS node $N$. The model’s input is bifurcated: the first input comprises the current observation of the leading and trailing vehicles in agent $j^{h}$'s lane and the adjacent ones. The subsequent input includes the planned trajectories of CAVs with higher priority than $j^{h}$, within a preview horizon $p^c$. These inputs are structured as graphs; specifically, the first input forms a graph whose nodes featurize the positions and velocities of the leading and following vehicles on its current and adjacent lane, denoted as $o(j^{h})$. This graph feeds directly into a GCN block. For the second input, we select CAVs that have higher priority than $j^{h}$ at current planning step, and encode their planned trajectories to a graph sequence denoted as $\{\tau(p) \mid p \prec_Nj^{h},\ p \in \boldsymbol{C}\}$ (line 31). Each graphs within this sequence has the future positions and velocities of CAVs that have higher priority than $j^{h}$ as node features. A second GCN block processes each graph in the sequence, employing shared weights and averaging the outcomes across the sequence's temporal dimension. We assume that each graph is fully connected.

Previous research suggests that the longitudinal behavior of highway drivers is relatively straightforward to predict \cite{MunigetyMathew2016}. Consequently, we posit that an HDV will sustain its current speed for a foreseeable interval, focusing our predictive efforts on the vehicle’s lateral movements. The combined GCN outputs are then channeled through two fully-connected layers, culminating in a sigmoid layer that computes the probability of the lane change decision of $j^{h}$ at next timestep. The outcome is a sequence of future lane change decision for $j^{h}$, which is combined with the library of motion primitives to generate the corresponding trajectory prediction $\tau(j^{h})$. 
\begin{figure}[!h]
\centering
\includegraphics[width=\columnwidth]{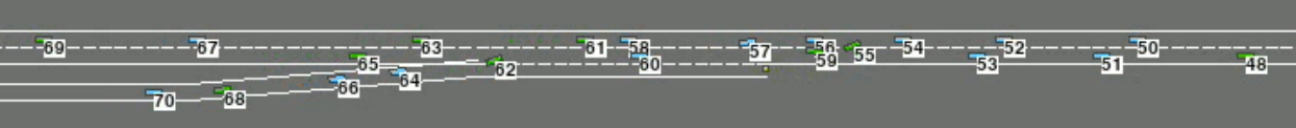}
\caption{Highway Simulator: CAVs are colored in green and HDVs are colored in blue}
\label{fig:higway_env}
\end{figure}

\begin{figure*}[!t]
\centering
\includegraphics[width=0.95\textwidth]{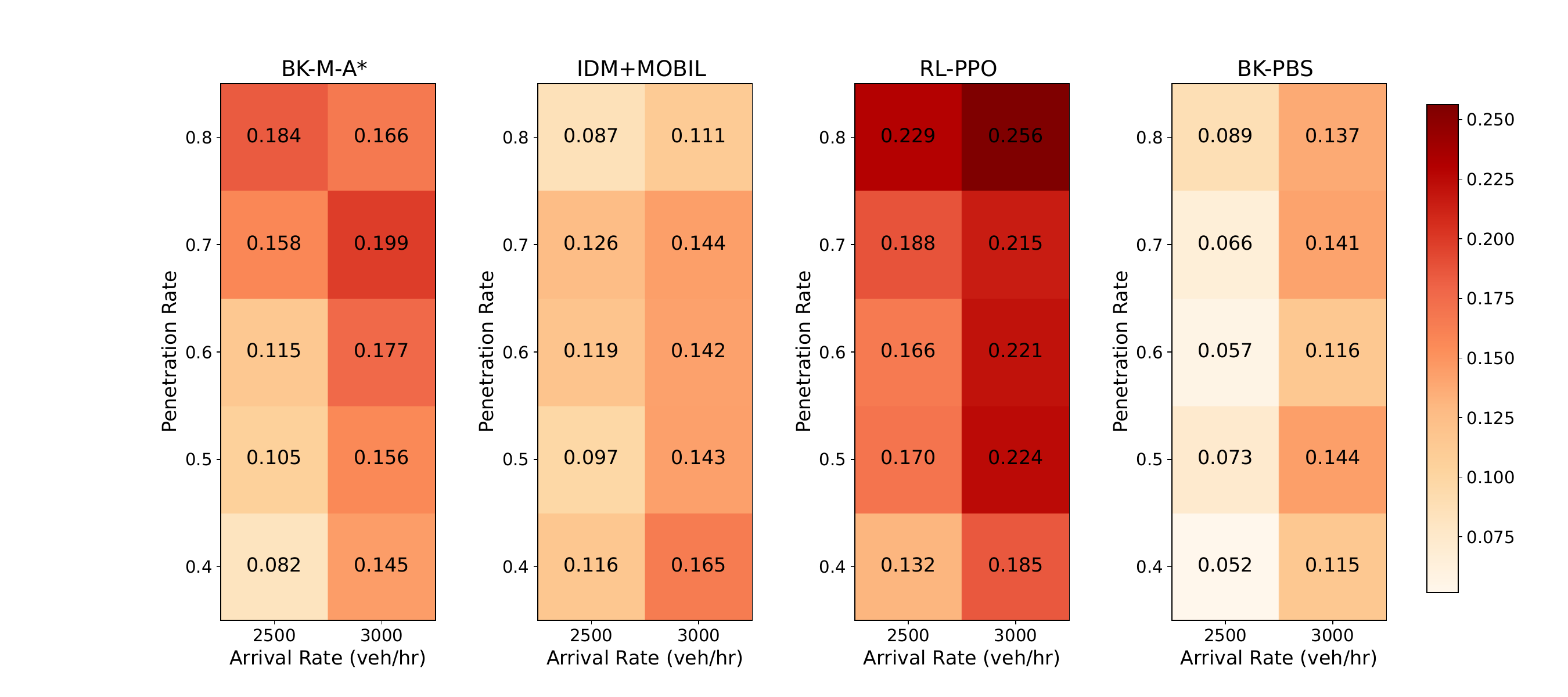}
\caption{Controllable collision rate under different vehicle arrival rates and CAV penetration rates}
\label{fig:collision}
\end{figure*}

\section{Experimental Setup}
\label{sec:setup}

\subsection{Simulation Environment}
\label{subsec: simulation env}

We modify HighwayEnv \cite{highway-env} to simulate the system with discretization $\text{d}t = 0.2$s for $H = 400$ timesteps. Fig.~\ref{fig:higway_env} illustrates the layout of the highway section. The penetration rate of CAVs is denoted as $\alpha$. The highway section has length $l=460$, with each lane has width $w_\text{lane} = 4.5$m and the merging zone length $l_m = 180$m.

We set arrival rate $\lambda$ veh/hr with initial speed sampled from a uniform distribution $v_0 \sim U(25, 35)$ m/s. The CAV penetration rate $\alpha$ is defined as the ratio of populated CAVs to the total number of populated vehicles. Each vehicle has length $\ell_k = 5$m and width $2$m. Vehicles collide when their bounding boxes overlap. Crashed vehicles are removed from the scene after a collision happens. The speed limit on the highway is set as $v_\text{max} = 35$m/s. The longitudinal behavior of HDVs is governed by the IDM \cite{IDM_2000} while the lateral behavior is governed by MOBIL model \cite{MOBIL1999}. We use the default parameters from \cite{highway-env} for the HDV driver models, except the target speed of IDM is sampled uniformly from 25 m/s to 35 m/s.

For a HDV driving on the ramp, its lateral behavior is governed by the stochastic merging model from \cite{Lin2024}, which shapes the probability of taking the merge action in proportional to the distance to the end of the merging zone.

\subsection{Prediction Model}
We train the prediction model used in BK-PBS and BK-M-A* using supervised learning. For data collection, we run a rollout policy for CAV control and record the states of every CAV and HDV at every simulation step. We post-process the data to match the input-output pairs for the prediction network described in \ref{subsec: GCN}, resulting a dataset which has 30K training samples in total.

The output from all hidden layers in the GCN and fully connected layers has size 64, expect for the last fully connected layers which outputs a single value to the sigmod function to determine the probability for lane change decision of a HDV. For the prediction network used in BK-M-A*, we do not include the planned trajectories of CAVs as conditional input, since there is no priority order among vehicles when planning under BK-M-A*. We train the model for 1000 epochs using Adam Optimizer \cite{Kingma2014AdamAM} with learning rate $lr = 0.001$ and batch size as 64. The training is perform on a desktop with a single RTX 3090 Ti GPU, and the total training time is around 4 hours.

Note that the model is trained using data collected from environment setting with certain vehicle arrival rate and CAV penetration rate, while deployed in different environment settings.

\subsection{Planning Algorithms}
\label{sub sec:planning Algorithms}
We compare BK-PBS with three baselines: BK-A*, IDM+MOBIL and RL-PPO. 

BK-M-A* is a decentralized planning algorithm that determines the trajectory of each CAV independently using M-A*. In BK-M-A*, a CAV under planning sees its surrounding vehicles as dynamics obstacles, whose trajectories are predicted using a learned prediction model and encoded as dynamic obstacle constraints into the M-A* planning of that CAV. Compared to BK-PBS, BK-M-A* does not have a high-level planning procedure to resolve the collision conflicts between vehicles. For M-A* used in both BK-PBS and BK-M-A*, according to \ref{subsec: MAPF Formulation} and \ref{subsec:A_star}, for CAVs driving on the main road, we set the goal distance in the first phase of A* planning as $d = 70$ m, while the revised goal distance in the second phase of A* planning as $d' = 30$ m.

IDM+MOBIL utilizes the rule-based car-following models \cite{IDM_2000, MOBIL1999} for CAV planning. We choose its parameters same as the parameters of the HDV driver model, expect that the target speed of IDM for CAV is set as 35 m/s.

Reinforcement learning is a commonly used approach to coordinate CAV in a highway driving scenario. RL-PPO uses Proximal Policy Optimization \cite{schulman2017proximal} to offline train a control policy for a CAV to select the motion primitives. The RL agent is trained in a single CAV environment with a reward function of maintaining a target speed and avoiding collision with up to 14 surrounding HDVs. The policy and value function networks both have two layers and hidden dimensions of 256. We run training with a learning rate of 0.001, discount factor of 0.95, batch size of 512, and 10 epochs per gradient update for 1e8 total environment steps (around 1 day on 48 CPU cores). After training, we inference the policy in a multi-agent environment described in \ref{subsec: simulation env}. The trained policy is applied to each CAV independently with weights sharing.  

We run all the planning algorithms in a rolling-horizon fashion, where the planning frequency is $5$ Hz. After each planning round, we take the first $n$ primitives from the each planned primitive sequence of a CAV. In all experiments, $n$ is set to 1. 

For robustness study we sweep $\alpha$ from 0.4 to 0.8 in both low density ($\lambda = 2500$ veh/hr) and high density ($\lambda = 3000$ veh/hr) traffic. We test each planning algorithm on $5$ environment seeds in all settings and average the results. We track the number of crashed vehicles and travel delay of every vehicle that successfully arrives at the end of the highway section without collision.

\section{Experimental Results}
\label{sec:results}

\subsection{Evaluation of Prediction Model}

We conduct an offline evaluation of the trained prediction model $b_{\theta}$, using a dataset procured in a same manner to the training dataset. As detailed in \ref{subsec: GCN}, this model predicts the lane-changing decisions of HDV with binary outcomes: 1 signifies executing a lane change, while 0 indicates maintaining the current lane. Due to the different lateral driver model for HDV on ramps versus main roads, we partition the evaluation dataset into two subsets. One subset contains samples exclusively from ramps while the other contains samples solely from main roads, allowing us to assess the model’s performance in predicting lane-changing decisions governed by different lateral driver models.

Each subset comprises 4000 samples, and we present the evaluation outcomes in Table.\ref{tab:confmat_ramp} for ramp data and Table.\ref{tab:confmat_main} for main road data.

\begin{table}[!h]
    \caption{\textbf{Lane-change prediction on the ramp}}
    \label{tab:confmat_ramp}
    \centering
    \begin{tabular}{c|cc}
      \toprule
        \bfseries accuracy = 92.6\% & \bfseries Predicted No & \bfseries Predicted Yes\\
        \midrule
        \bfseries Actual No  &2469& 88 \\
        \bfseries Actual Yes  &205&1238\\
        
        \bottomrule
    \end{tabular}\label{tab:configs}
\end{table}

\begin{table}[!h]
    \caption{\textbf{Lane-change prediction on the main road}}
    \label{tab:confmat_main}
    \centering
    \begin{tabular}{c|cc}
      \toprule
        \bfseries accuracy = 94.1\% & \bfseries Predicted No & \bfseries Predicted Yes\\
        \midrule
        \bfseries Actual No  &3237& 57 \\
        \bfseries Actual Yes  &181&525\\
        
        \bottomrule
    \end{tabular}\label{tab:configs}
\end{table}

We find the trained model $b_{\theta}$ attains high accuracy in overall lane change prediction. However, a marked disparity is evident between the ratio of true negatives to true positives in predictions of lane-changing on main roads compared to ramps. This discrepancy suggests that the model's proficiency in predicting HDV lane changes on main roads is somewhat limited, a focal point that may necessitate further model refinement to enhance predictive accuracy.

\subsection{Controllable Collision Rate}

After each simulation episode, we compute the collision rate as the ratio between the crashed vehicles and total number of vehicles that inserted to the scene. Note that a HDV can crashed into another HDV during simulation (this happens naturally within our simulator), which is out of the algorithm's control; thus we only focus on collisions that have a CAV involved, and we named the resulting collision rate as the controllable collision rate.

We report the controllable collision rate in Fig.\ref{fig:collision}. A visual inspection reveals that the proposed method BK-PBS achieves relatively low collision rate overall, as indicated by the predominance of lighter shades. As mentioned in \ref{sub sec:planning Algorithms}, the values shown are averaged over 5 random seeds for each scenario setting.

RL-PPO exhibits the highest collision rates across all penetration and arrival rates. Notably, its collision rates escalate with increasing vehicle arrival rates, suggesting that RL-PPO struggles to maintain safety in denser traffic conditions. IDM+MOBIL appears to fare better, with its heat map exhibiting consistently lighter shades. It exhibits a lesser dependency on arrival rate changes, which implies the robustness of these rule-based driver models in diverse traffic densities. We found collision rates with BK-PBS tend to rise alongside arrival rates, and such trend can also be found with BK-M-A*.

BK-M-A* shows intermediate performance but demonstrates elevated collision rates under all settings compared to BK-PBS, which is due to the fact that BK-M-A* does not resolve conflicts between vehicles while a CAV-CAV conflict can be effectively solved by BK-PBS, resulting in less collisions between two CAVs. This explanation can be validated by observing that the collision rate with BK-M-A* grows quickly as penetration rate increases.

One interesting observation is that at high traffic densities ($\lambda=3000$ veh/hr), IDM+MOBIL occasionally records fewer collisions compared to BK-PBS. This could be because, in heavy traffic, the MOBIL model restricts frequent lane changes due to the presence of vehicles in adjacent lanes most of the time. In contrast, BK-PBS consistently seeks the shortest path to the destination, resulting in more frequent lane shifts and side collisions.

\subsection{Travel Delay}

We compute the average delay of vehicles that successfully arrives at the end of the highway section without collision according to Eq.\ref{eq:delay} and report it in Table.\ref{tab:delay low} and Table.\ref{tab:delay high}.

\begin{table}[!h]
    \caption{\textbf{Average delay (sec) at $\lambda = 2500$ veh/hr }}
    \label{tab:delay low}
    \centering
    \begin{tabular}{cccccc}
      \toprule
        \bfseries $\alpha$ & \bfseries 0.4 & \bfseries 0.5 & \bfseries 0.6 & \bfseries 0.7 & \bfseries 0.8\\
        \midrule
        \bfseries BK-M-A* &6.24 & \textbf{5.79} & 5.63 & 5.50 & 5.67\\
        \bfseries IDM+MOBIL &\textbf{5.82} &6.05  & 5.72 & 5.83 & 5.59\\
        \bfseries RL-PPO &7.19 &7.54  & 7.43 & 7.30&7.09\\
        \bfseries BK-PBS &6.51& 6.31& \textbf{5.61} & \textbf{5.45} & \textbf{5.51}\\
        
        \bottomrule
    \end{tabular}\label{tab:configs}
\end{table}

\begin{table}[!h]
    \caption{\textbf{Average delay (sec) at $\lambda = 3000$ veh/hr}}
    \label{tab:delay high}
    \centering
    \begin{tabular}{cccccc}
      \toprule
        \bfseries  $\alpha$ & \bfseries 0.4 & \bfseries 0.5 & \bfseries 0.6 & \bfseries 0.7 & \bfseries 0.8\\
        \midrule
        \bfseries BK-M-A* &\textbf{6.81} &6.60 & \textbf{5.78} & 6.03 & 6.01\\
        \bfseries IDM+MOBIL &7.46 &7.72  & 7.19 & 7.16 & 7.98\\
        \bfseries RL-PPO &8.11 &8.13  & 8.09 & 7.74 &7.56\\
        \bfseries BK-PBS &7.02& \textbf{6.50} & 6.06 & \textbf{5.77} & \textbf{5.80}\\
        
        \bottomrule
    \end{tabular}\label{tab:configs}
\end{table}

While an satisfying outcome would be for BK-PBS to facilitate smoother traffic flow and yield the lowest travel delay relative to the three benchmark algorithms, it appears that under certain conditions, BK-M-A* and IDM+MOBIL exhibit the lowest travel delays.

This result may seem misleading, as in scenarios where a baseline surpasses BK-PBS in travel delay, it concurrently incurs a substantially higher collision rate, as depicted in Fig.\ref{fig:collision}. During simulation, crashed vehicles are removed from the scene after a collision happens. This leads to less congested traffic conditions and, consequently, enabling remaining vehicles to accelerate, which artificially reduces travel delay.

A more equitable method of comparison would consider travel delays at comparable collision rates. When specifically comparing IDM+MOBIL and BK-PBS at an arrival rate of $\lambda = 3000$ veh/hr and accounting for similar collision rates, it becomes evident that BK-PBS excels in dense traffic conditions. We observe that BK-PBS significantly reduces total travel delay by 15\%-20\% compared to IDM+MOBIL in dense traffic, reaffirming its effectiveness in smoothing traffic flow.


\section{Conclusions and Future Work}

This paper presents the Behavior Prediction Kinematic Priority Based Search (BK-PBS), a novel framework that utilizes an offline-trained conditional prediction model for forecasting the reactions of HDVs to CAV actions. This prediction model is adeptly integrated into Priority Based Search (PBS), which employing a low-level multi-phase A* search with motion primitives that respect kinematic constraints. In simulated highway merging scenarios with varied traffic densities and AV penetrations, BK-PBS outperformed existing rule-based and RL algorithms, reducing collisions and travel delays. This method holds promise for improving the coordination between humans and robots in mixed-traffic environments, for example, cooperative navigation of mobile robots among crowds. 

We aim to evolve the BK-PBS framework further by integrating more sophisticated behavior prediction methodologies, enhancing safety measures with the implementation of safety envelopes like control barrier functions, and extending its applicability to larger and more general networks. Additionally, since crashed vehicles are removed from the scene in our simulation, validating BK-PBS in physical testbeds that account for replanning in response to chain reactions caused by crashes is a crucial next step. This will bring us closer to realizing its potential in real-world systems.


\bibliographystyle{ieeetr}
\bibliography{references}

\end{document}